\documentclass{article}

\usepackage{PRIMEarxiv}

\usepackage[utf8]{inputenc} 
\usepackage[T1]{fontenc}    
\usepackage{hyperref}       
\usepackage{url}            
\usepackage{booktabs}       
\usepackage{amsfonts}       
\usepackage{nicefrac}       
\usepackage{microtype}      
\usepackage{lipsum}
\usepackage{fancyhdr}       
\usepackage{graphicx}       
\graphicspath{{media/}}     

\title{InvertibleNetworks.jl: A Julia package for scalable normalizing flows
}

\author{
  Rafael Orozco \\
  Georgia Institute of Technology \\
  \texttt{rorozco@gatech.edu} \\
   \And
   Philipp Witte \\
  Microsoft Research \\
    \And
   Mathias Louboutin \\
  Devito Codes\\
    \And
   Ali Siahkoohi \\
  Rice University \\
    \And
   Gabrio Rizzuti \\
  Shearwater GeoServices \\
    \And
   Bas Peters \\
  Computational Geosciences Inc \\
   \And
    Felix J. Herrmann \\
  Georgia Institute of Technology \\
}

\begin{document}
\maketitle



\section{Summary}
Normalizing flows is a density estimation method that provides efficient exact likelihood estimation and sampling \cite{dinh2014nice} from high dimensional distributions. This method depends on the use of the change of variables formula which requires an invertible transform. Thus normalizing flow architectures are built to be invertible by design \cite{dinh2014nice}. In theory, the invertibility of architectures constrains the expressiveness but the use of coupling layers allows normalizing flows to exploit the power of arbitrary neural networks that need not be invertible \cite{dinh2016density} and layer invertibility means if properly implemented many layers can be stacked to increase expressiveness without creating a training memory bottleneck.  

The package we present, InvertibleNetworks.jl, is a pure Julia \cite{bezanson2012julia} implementation of normalizing flows. We have implemented many relevant neural network layers, including GLOW 1x1 invertible convolutions \cite{kingma2018glow}, affine/additive coupling layers \cite{dinh2014nice}, Haar wavelet multiscale transforms \cite{haar1909theorie} and Hierarchical invertible neural transport (HINT) \cite{kruse2021hint} among others. These modular layers are easily composed and modified to create different types of normalizing flows. As starting points, we have implemented RealNVP, GLOW, HINT, Hyperbolic networks \cite{lensink2022fully} and their conditional counterparts for users to quickly implement their individual applications. 

\section{ Statement of need}
This software package focuses on memory efficiency. The promise of neural networks is in learning high-dimensional distributions from examples thus normalizing flow packages should allow easy application to large dimensional inputs such as images or 3D volumes. Interestingly, the invertibility of normalizing flows naturally alleviates memory concerns since intermediate networks activations can be recomputed instead of saved in memory, greatly reducing the memory needed during backpropagation. The problem is that directly implementing normalizing flows in automatic differentiation frameworks such as PyTorch \cite{paszke2017automatic} will not automatically exploit this invertibility. The available packages for normalizing flows such as: nflows \cite{nflows}, normflows \cite{stimper2023normflows} and FrEIA \cite{freia} are built depending on automatic differentiation frameworks thus do not exploit invertibility for memory efficiently. 

\section{Memory efficiency}
By implementing gradients by hand instead of depending completely on automatic differentiation, our layers are capable of scaling to large inputs. By scaling, we mean that these codes are not prone to out-of-memory errors when training on GPU accelerators. Indeed, previous literature has described memory problems when using normalizing flows as their invertibility requires the latent code to maintain the same dimensionality as the input \cite{khorashadizadeh2023conditional}.

\begin{figure}
  \centering
\includegraphics[width=0.9\textwidth]{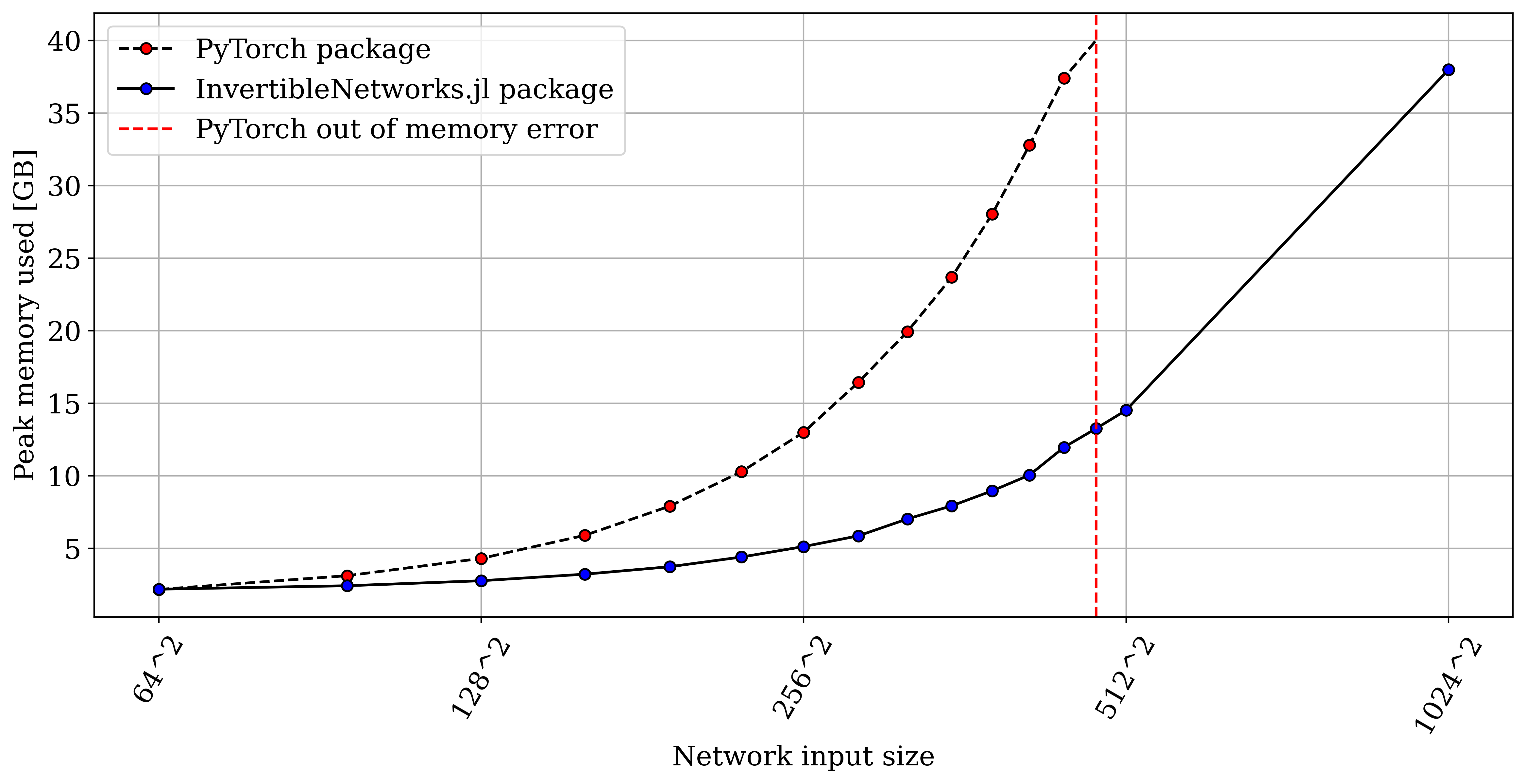}
  \caption{Our package InvertibleNetworks.jl provides memory frugal implementations of normalizing flows. Here, we compare our implementation of GLOW with an equivalent implementation in a PyTorch package.  Using a 40GB A100 GPU, the PyTorch package can not train on image sizes larger than 480x480,  while our package can handle sizes larger than 1024x1024.}
  \label{fig:memory}
\end{figure}

In Figure \ref{fig:memory}, we show the relation between input size and the memory required for a gradient calculation in a PyTorch normalizing flow package (normflows \cite{stimper2023normflows}) as compared to our package. The two tests were run with identical normalizing flow architectures. We note that the PyTorch implementation quickly increases the memory load and throws an out of memory error on the 40GB A100 GPU at the spatial image size of 480x480 while our InvertibleNetworks.jl implementation still has not run out of memory at spatial size 1024x1024. Note that this is in the context of a typical learning routine, so the images include 3 channels (RGB) and we used a batchsize of 8.

\begin{figure}
  \centering
\includegraphics[width=0.9\textwidth]{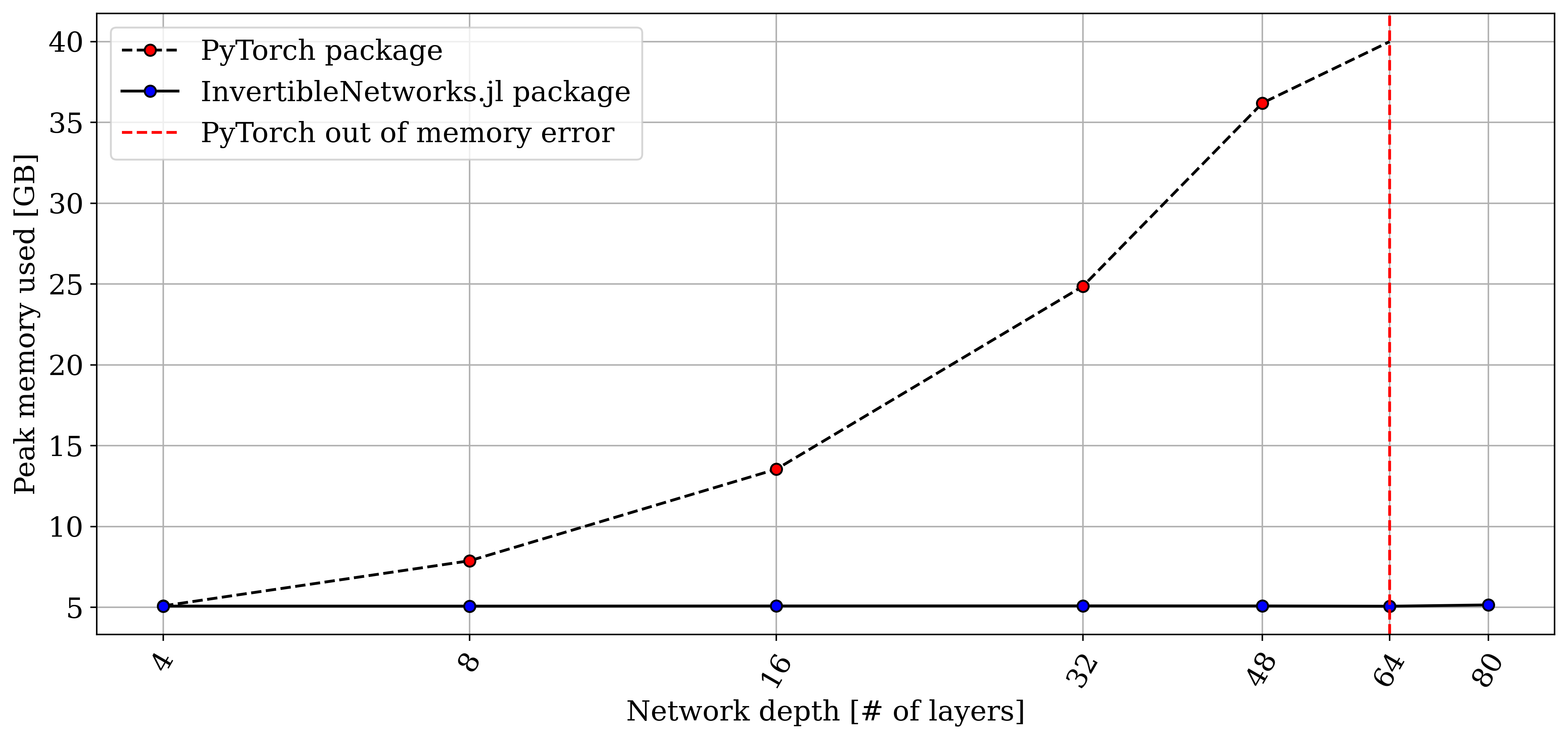}
  \caption{Due to the invertibility of the normalizing flow architecture, the memory consumption does not increase as we increase the depth of the network. Our package properly exploits this property thus shows constant memory consumption whereas the PyTorch package does not.}
  \label{fig:memory-depth}
\end{figure}

Since traditional normalizing flow architectures need to be invertible these might be less expressive as compared to non-invertible counterparts. In order to increase their expressiveness, practitioners stack many invertible layers to increase the overall expressive power. Increasing the depth of a neural network would in most cases increase the memory consumption of the network but in this case since normalizing flows are invertible, the memory consumption does not increase. Our package displays this phenomena as shown in Figure \ref{fig:memory-depth} while the PyTorch (normflows) package that has been implemented with automatic differentiation does not display this constant memory phenomena.

\section{Ease of use}
Although the normalizing flow layers gradients are hand-written, the package is fully compatibly with ChainRules \cite{frames_white_2023_10100624} in order to integrate with automatic differentiation frameworks in Julia such as Zygote \cite{innes2019differentiable}. This integration allows users to add arbitrary neural networks which will be differentiated by automatic differentiation while the memory bottleneck created by normalizing flow gradients will be dealt with InvertibleNetworks.jl. The typical use case for this combination is the summary networks used in amortized variational inference such as BayesFlow \cite{radev2020bayesflow} which has been implemented in our package. 

All implemented layers are tested for invertibility and correctness of their gradients with continuous integration testing via GitHub actions.  There are many example for layers, networks and  application workflows allowing new users to quickly build networks for a variety of applications. The ease of use is demonstrated by the publications that made use of the package.

Many publications have used InvertibleNetworks.jl for diverse applications including: change point detection, \cite{peters2022point}, acoustic data denoising \cite{kumar2021enabling}, seismic imaging \cite{rizzuti2020parameterizing, siahkoohi2021preconditioned,siahkoohi2022wave,siahkoohi2023reliable,louboutin2023learned,alemohammad2023self}, fluid flow dynamics \cite{yin2023solving}, medical imaging \cite{orozco2023adjoint, orozco2023amortized,orozco2021photoacoustic, orozco2023refining} and monitoring CO2 for combating climate change \cite{gahlot2023inference}.

\section{Future work}
The neural network primitives (convolutions, non-linearities, pooling etc) are implemented in NNlib.jl abstractions thus support for AMD, Intel and Apple GPU can be trivially extended. Also, while our package currently can handle 3D inputs and has been used on large volume-based medical imaging \cite{orozco2022memory} there are interesting avenues of research regarding the "channel explosion" seen in invertible down and upsampling used in invertible networks \cite{peters2019symmetric}.

\bibliographystyle{unsrt}  
\bibliography{references}

\end{document}